\pdfoutput=1
\documentclass[11pt]{article}

\usepackage[preprint]{acl}

\usepackage{times}
\usepackage{latexsym}

\usepackage[T1]{fontenc}
\usepackage[utf8]{inputenc}
\usepackage{microtype}
\usepackage{inconsolata}

\usepackage{graphicx}
\usepackage{multirow}
\usepackage{enumitem}
\usepackage{arydshln}
\usepackage[most]{tcolorbox}

\definecolor{highlight}{RGB}{255, 230, 204}
\definecolor{soft}{RGB}{240, 248, 255}

\title{Pairing Analogy-Augmented Generation with Procedural Memory for Procedural Q\&A}

\author{
 \textbf{K Roth\textsuperscript{1}\thanks{Equal contribution.}},
 \textbf{Rushil Gupta\textsuperscript{1}\footnotemark[1]},
 \textbf{Simon Halle\textsuperscript{2}},
 \textbf{Bang Liu\textsuperscript{1,3}\thanks{Corresponding author.}}
\\
\\
 \textsuperscript{1}Universit\'e de Montr\'eal \& Mila,
 \textsuperscript{2}Thales Canada,
 \textsuperscript{3}Canada CIFAR AI Chair
\\
\{kyle.roth, rushil.gupta, bang.liu\}@umontreal.ca \\
simon.halle@thalesgroup.com
}

\begin{document}
\maketitle
\begin{abstract}
Large language models struggle to synthesize disparate pieces of information into a coherent plan when approaching a complex procedural task. In this work, we introduce a novel formalism and structure for such procedural knowledge. Based on this formalism, we present a novel procedural knowledge dataset called LCStep, which we created from LangChain tutorials. To leverage this procedural knowledge to solve new tasks, we propose analogy-augmented generation (AAG), which draws inspiration from the human ability to assimilate past experiences to solve unfamiliar problems. AAG uses a custom procedure memory store to retrieve and adapt specialized domain knowledge to answer new procedural tasks. We demonstrate that AAG outperforms few-shot and RAG baselines on LCStep, RecipeNLG, and CHAMP datasets under a pairwise LLM-based evaluation, corroborated by human evaluation in the case of RecipeNLG.
\end{abstract}
\section{Introduction}
\noindent Procedural knowledge, or ``knowledge-how'', is a form of knowledge that involves the ability to \textit{do} something, as opposed to propositional knowledge, which is the awareness of a fact (``knowledge-that'') \cite{sep-knowledge-how}. A hallmark of procedural knowledge is that the agent is expected not only to understand the facts but also \textit{plan} and \textit{reason} about actions in its environment.

The community has made efforts both theoretically \cite{old-procedural-knowledge} and empirically via LLMs to enable information systems that explicitly handle procedural knowledge in real-world \cite{wang2024llmbasedrobot} and virtual environments~\cite{llmsZeroShotPlanners,wang2024agent-survey,tan2024towards, brahman2023plasma}. However, these works have primarily exploited prompt engineering to leverage the model's internal or limited external knowledge without taking inspiration from human cognition, which excels in these tasks.

LLMs consistently struggle to perform well in specialized domains not well represented in training data and require up-to-date knowledge \cite{kandpal2023large}, usually added through fine-tuning or as augmented external memory. On the contrary, humans demonstrate \textit{analogical reasoning} abilities in these situations \cite{Parsons2022-PARTNC}, adapting relevant knowledge from their past experiences in new ways to achieve a goal.

\begin{figure}[t]
\centering
  \includegraphics[width=0.5\textwidth]{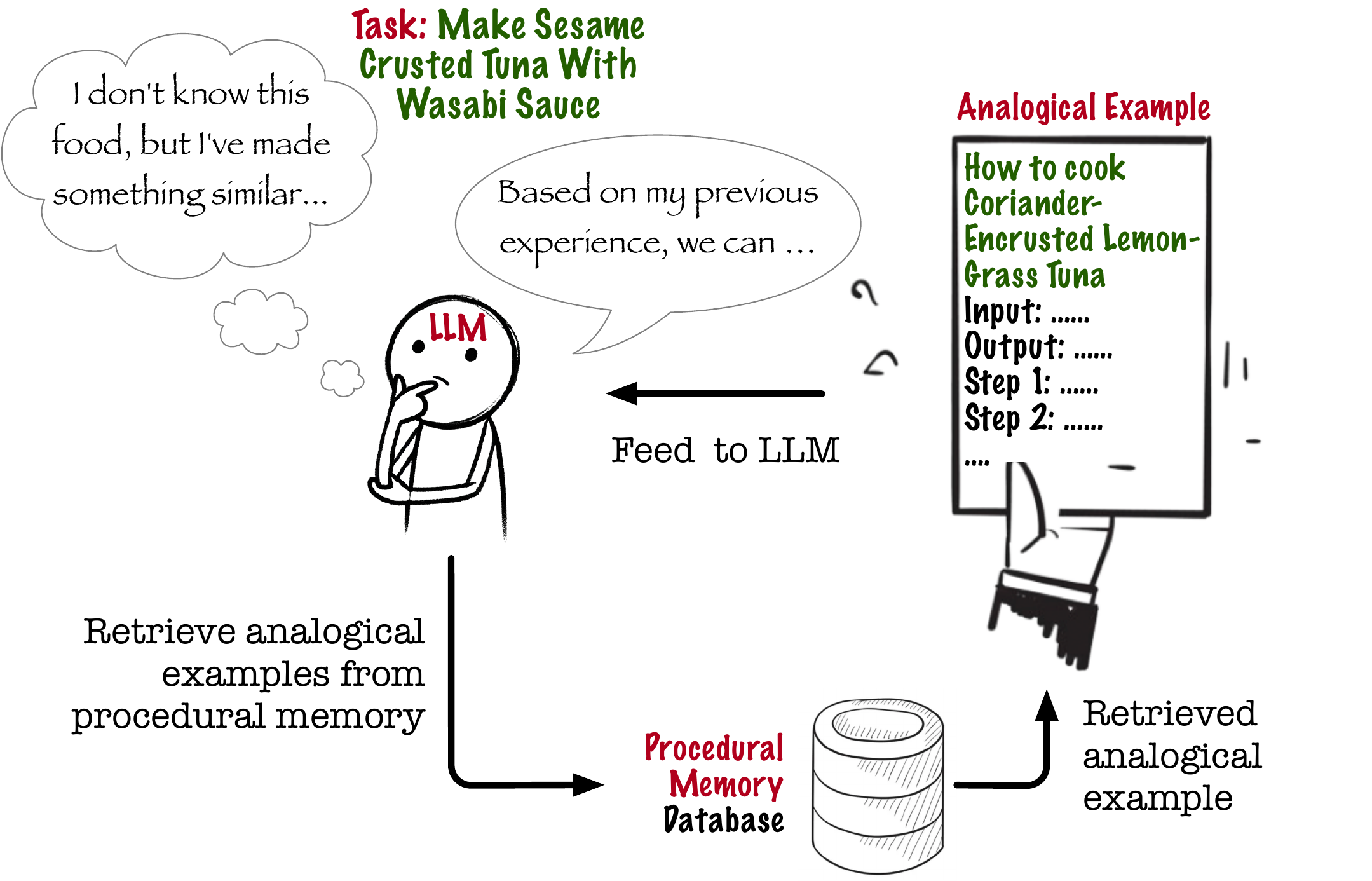}
  \caption {An illustration of using the AAG system to generate a recipe for an unseen food.}
  \label{fig:cartoon}
\end{figure}

Analogy is a central ability of human cognition and has been formally theorized by human psychologists, the most common being \textit{Structure Mapping Theory} \cite{Structure-Mapping-Theory}. The theory defines analogy as mapping knowledge from one situation (base) into another (target). According to the theory, a situation consists of objects, their properties, and the \textit{relationships} that characterize their interactions. It is not \textit{object attributes} that gets mapped in an analogy; it is the similarity in the relationships between objects that plays the key role \cite{GENTNER2012130}. The \textit{systematicity principle}, a crucial part of the theory, emphasizes that people prefer to map predicates with higher-order relations than isolated predicates.

To ground this, consider the recipe example presented in Figure~\ref{fig:cartoon}. Each recipe forms a situation, with all the different ingredients and intermediate dishes constituting the objects. The relations in the situation correspond to the cooking operations, which are complex predicates involving two or more arguments. For example, consider the step, ``dip the tuna in a crusting batter infused with coriander'', which could be thought of as the predicate: \texttt{Dip(tuna, Batter(Coriander, \dots))}, comprising one simple and one complex predicate as an argument. The LLM agent possesses knowledge of such cooking operations from examples in its memory and leverages them in the new question on sesame-crusted tuna, which would involve a similar predicate: \texttt{Dip(tuna, Batter(Sesame, \dots)}.

Inspired by human analogical reasoning, we propose Analogy-Augmented Generation \textbf{(AAG)}, a novel system for procedural question answering based on the \textit{Rewrite-Retrieve-Read} framework \cite{ma2023query}. The AAG pipeline retrieves analogical examples from a specialized procedural memory in a multi-step process, creating a procedure that accomplishes the query (see Figure~\ref{fig:model-diagram}).\\
Our contributions in this paper are the following:
\begin{itemize}[leftmargin=*] \vspace{-3mm}
    \item We provide an easily extensible formalism for text-based procedural knowledge that plugs easily into existing information retrieval systems.
    \item We describe a novel dataset, \textbf{LCStep}, consisting of procedures to do tasks with the LangChain Python library \cite{LangChain}. The recency of this data presents an opportunity to test procedural knowledge systems on a topic not present in some LLMs' training data.
    \item We propose a novel LLM-based system called Analogy-Augmented Generation (AAG) that capitalizes on the above-defined formalism for procedural knowledge and implements a workflow inspired by human analogical reasoning.
    \item We evaluate this system on an unfamiliar domain using our LCStep dataset and again on a familiar domain using the RecipeNLG dataset~\cite{bien-etal-2020-recipenlg}. We ablate each separable component of the AAG system to measure the contribution of each. We also stretch our formal representation and system by applying it to CHAMP \cite{mao2024champ}, a challenging math problem dataset. Through LLM-based pairwise evaluation and a blind human study, we conclude that the AAG system generates more coherent and plausible procedures for the input queries than existing baselines, including RAG.
\end{itemize}

\section{Procedure Generation Task}
\label{sec:task}

We define a procedure as $(x,y,(s_1,\ldots,s_k))$ where:

\begin{itemize}[leftmargin=*]\vspace{-2mm}
    \item $x$ is a string containing the available inputs,\vspace{-2mm}
    \item $y$ is the string containing the output of the procedure, and\vspace{-2mm}
    \item $(s_1,\ldots,s_k)$ is a sequence of ordered steps, each of which is a string.\vspace{-1mm}
\end{itemize}

\noindent This structure can be seen as a linear directed chain graph starting with the input as the root, the output as the leaf node, and each step $s_i$ as an intermediate node in the chain.

Unlike some existing conceptions of procedural knowledge~\cite{old-procedural-knowledge}, our formalism depends on the text of the steps to describe any non-linear control flow. The sequential format makes it easier for LLMs to grasp task-relevant patterns, fostering analogical reasoning and the reuse of experience to tackle new tasks.
Our approach retains composability: two procedures $(x,y,(s_1,\ldots,s_k))$ and $(y,z,(t_1,\ldots,t_\ell))$ can be combined into $(x,z,(s_1,\ldots,s_k,t_1,\ldots,t_\ell))$. This supports the generation of $(s_1,\ldots,s_k)$ from $x$ and $y$, for example, generating recipe steps from a title and ingredient list.

Procedural knowledge differs from knowledge graphs, as it captures skill-like sequences that can be directly translated into agent actions or code APIs, emphasizing "how" to achieve outcomes. Functionally, this structure is an improvement over plain text because it reduces noise and isolates essential procedural information, improving data quality for enhancing LLMs' reasoning and planning.
This paper focuses on linear chains due to the complexity of dependency extraction from raw procedure data, but our future work will focus on expanding this to explicitly represent dependencies within procedures.

Under this framework, the task of procedure generation is to generate $(s_1,\ldots,s_k)$ given $x$ and $y$. For example, a procedure generation system for food recipes will be expected to generate the cooking steps given the recipe title and ingredients.

\section{LCStep Dataset}
\label{sec:lcstep}

A key goal in this work is to test systems' domain adaptation ability to answer procedural knowledge questions for domains not present or not up to date in LLM training data. Across a wide variety of tasks, the expense of model training makes it infeasible to continually retrain as new domains become relevant or as information is updated~\cite{gao2024retrievalaugmented}. Retrieval-augmented generation (RAG) is popularly used to leverage external data to keep information up to date, but we expect our procedural knowledge system to provide further benefits, especially because answering procedural knowledge questions often requires information from disparate document sources.

For our experiments, we collected a dataset of procedural knowledge of the LangChain Python library, unseen by many extant LLMs. We selected LangChain as the domain for our dataset because it was published in 2022, which is later than the knowledge cutoff date for many web-scale LLMs, including GPT-3.5~\cite{openai2024gpt35turbo}, while also having plenty of documentation due to its popularity. Of course, any domain of public interest will quickly become incorporated into the training data for both commercial and academic models. Hence, a dataset of this nature has a limited time window in which to truly test domain adaptation. In this paper, we perform our experiments using a model with a cutoff date before 2022. This allows us to measure the extent to which systems can leverage the supplied procedural knowledge store to solve tasks in an unseen domain.

The LCStep dataset was collected from 180 tutorial pages in the Python section of the LangChain website. We used an LLM-enabled pipeline with human oversight and quality review to extract 276 procedures from these tutorials, representing each procedure in the formal structure defined in Section~\ref{sec:task}. We include details on the collection pipeline in Appendix~\ref{app:lcstep}.
\section{Our Method}
\label{sec:method}

\begin{figure*}[!ht]
\centering
  \includegraphics[width=0.75\linewidth]{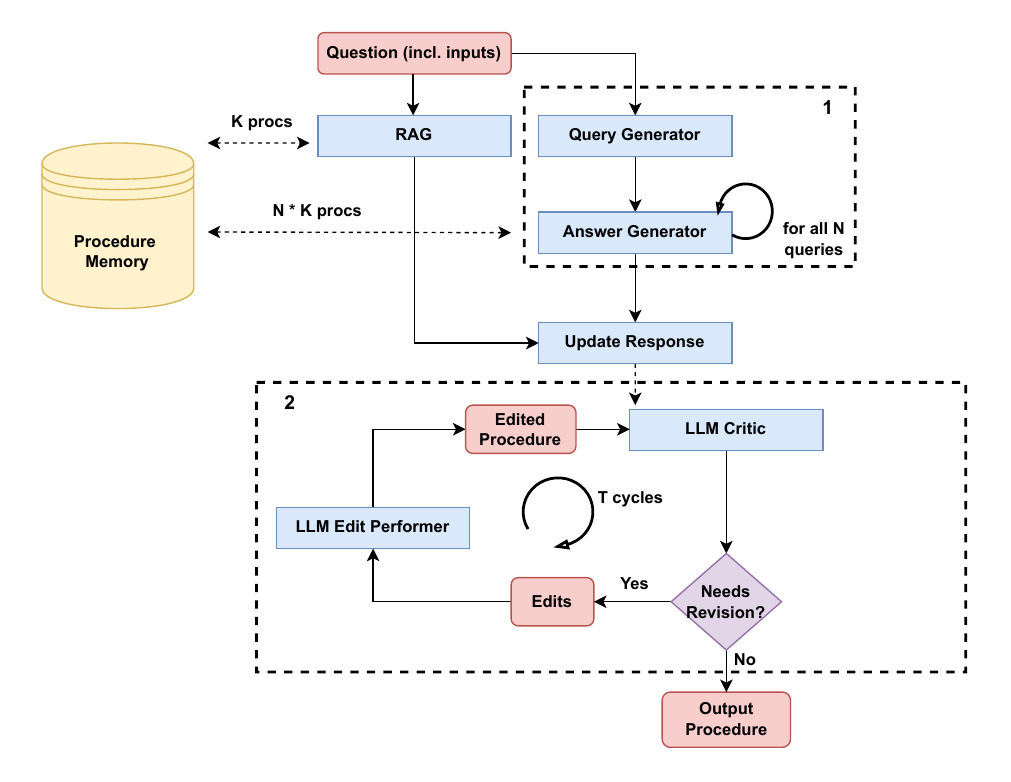}
  \caption {Schematic view of the AAG architecture. The input and intermediate outputs are enclosed in red rounded rectangles, whereas all modules involving an LLM are enclosed in blue rectangles. Boxes 1 and 2 denote our architectural contributions, where 1 shows the query and answer generation step, and 2 shows the iterative refinement with the self-critic step. The procedural memory store holds objects structured as procedures (see Section~\ref{sec-memory-store}). The RAG and answer generator modules interact with the memory store to retrieve $K$ and $N\times K$ procedures relevant to the input query and all the generated questions (one at a time), respectively.}
  \label{fig:model-diagram}
\end{figure*}

We propose an Analogy-Augmented Generation (AAG) system to tackle the task of procedural Q\&A. Our system comprises three key components: 1) \textbf{procedural memory store:} a retrieval database for procedures; 2) \textbf{query and answer generator}: the analogy step which breaks the input query into predicates (steps) and their corresponding questions and retrieves analogous procedures from memory to answer them; and 3) \textbf{iterative self-refinement}: an LLM-based critic which evaluates the candidate response, suggests edits, and then performs them. Figure~\ref{fig:model-diagram} shows a flow diagram of our system. We now describe each module in more detail, followed by an end-to-end description of the overall control flow. Please refer to Appendix~\ref{sec:prompts} for the full prompts used in each module.

\subsection{Procedural Memory}
\label{sec-memory-store}

The procedural memory store embeds whole procedures by formatting the linear graph in a text template natural to the domain. Every search to the memory store finds and returns $K$ nearest procedures to the search query in the embedding space. Practically, the store maintains the input, output, and steps as Python \texttt{str}, \texttt{str}, and \texttt{list[str]} types, respectively, when the procedures are retrieved, generated, or communicated from one module to another.

\subsection{Query Generator}
The input query to the AAG system comprises the goal (the procedure output $y$) and the resources we have access to (the procedure input $x$). Using this input query directly to search the memory can yield insufficient or noisy knowledge when the examples in the memory are indirectly related or don't use words similar to words in the query. Humans solve this problem by analogy, developing high-level mental maps of steps to achieve the goal and consequently linking and assimilating knowledge from relevant past experiences to compose the solution \cite{piaget1968genetic, vosniadou1989similarity}.

To approximate this in our system, we first prompt the LLM to generate a high-level outline of the steps in the required procedure (similar to HyDE \cite{gao2022precise}), followed by asking it to formulate questions for them that can be answered through analogical examples in the memory. We adapt the query rewriting prompt used by Ma et al.~(\citeyear{ma2023query}) for our task and generate a maximum of $N=4$ queries for a given input query. For example, given the input query: `\textit{create a custom LLM agent that can interact with users in a conversational manner using a specific language style using an LLM, Serp API}', the following four questions are generated by the query generator:
\begin{itemize}[leftmargin=*]\vspace{-1mm}
    \item How to train a language model for a specific language style?\vspace{-2mm}
    \item Best practices for integrating a language model into a conversational interface?\vspace{-2mm}
    \item How to use the Serp API to extract relevant information from search engine results?\vspace{-2mm}
    \item Strategy for optimizing the performance of a custom LLM agent in conversational interactions?\vspace{-1mm}
\end{itemize}

\subsection{Answer Generator}
Each question received from the query generator module is individually used to retrieve from the procedure memory. The challenge, however, is that the information required for the question is often implicit or hidden in the retrieved procedures.

Therefore, the answer generator module acts as a filter and prompts the LLM with the question and the retrieved procedures, asking it to generate a plausible answer by borrowing knowledge from the retrieved procedures. Note that this module relies heavily on LLM's ability to understand the semantics of the retrieved procedures as it filters out helpful content from the retrieved procedures. Finally, each question, along with its generated answer, is stacked together as \texttt{"Q: <Question> [newline] A: <Answer>"} to build a high-quality context, which is used by both the Update Response and the Edit performer module.

\subsection{Iterative Refinement with Self-Critic}
\label{sec:critic}
LLMs have been shown to act as powerful self-critics for their own outputs, evaluating them across several dimensions and suggesting feedback crucial to improving them \cite{madaans2023selfrefine, saunders2022selfcritiquing}. We augment the AAG system with an iterative refinement procedure guided by the same LLM-based critic for its output. As seen in Figure~\ref{fig:model-diagram}, the critic takes the output of the Update Response module, evaluates if any edits are required, and suggests them as a bulleted list. If no edits are needed, the critic appends the string \textit{`NO UPDATE REQUIRED'} to its response. The LLM then performs the suggested edits with a prompt that includes the list of edits, context from the answer generator, and the current version of the output procedure steps. The LLM critic then re-evaluates this edited version, and the cycle continues for a maximum of $T=3$ cycles. The output at the end of the iterative refinement is the final answer of the AAG system to the input query.

\subsection{End-to-End Workflow}
\label{sec:overall-workflow}
The AAG system takes as input a query by the user and outputs a bulleted list of steps that will lead the user to the desired goal. Using the terminologies defined in Section~\ref{sec:task}, the input query is of the format ``\textit{<output>} using \textit{<input>}'', where the user specifies the goal and the resources available to use to reach the goal. Assuming we have a procedural memory instantiated with the procedure objects as outlined in Sections~\ref{sec-memory-store} and \ref{sec:design-choice}, the detailed workflow of the AAG system is as follows (refer to Figure~\ref{fig:model-diagram} and the full trace in appendix~\ref{sec:prompts}):
\begin{enumerate}[leftmargin=*]\vspace{-2mm}
    \item The RAG module searches the procedural memory using the input query and retrieves the top $K=3$ relevant procedures (highest cosine similarity in the embedding space) from memory.\vspace{-2mm}
    \item The query generator module takes the input query and breaks it into $N=4$ queries about different steps of the solution.\vspace{-2mm}
    \item For each query, the answer generator module searches the memory and utilizes the retrieved procedures to generate an answer. All questions and corresponding answers are stacked together to form the context for downstream modules.\vspace{-2mm}
    \item The update response module is the first to update the RAG response using the above context and passes the updated response to the critic.\vspace{-2mm}
    \item The LLM critic decides if the response requires edits and suggests them in a bulleted list. \vspace{-2mm}
    \item The LLM edits its response using the suggestions given by the critic (same LLM) and the context from the answer generator module. \vspace{-2mm}
    \item The edited procedure is then re-evaluated for edits. Repeat step 5 for a maximum of $T=3$ cycles. The self-refined version is then returned as the output of the system.\vspace{-2mm}
\end{enumerate}

\section{Experiments}

We evaluate the effectiveness of the AAG system for procedure generation in three scenarios: an unseen domain (LCStep), a familiar domain (RecipeNLG), and CHAMP, a challenging domain for our formalism. We begin by describing the baselines used for all experiments, followed by discussing the evaluation methodology and design choices of the AAG system. We then examine the two case studies of LCStep and RecipeNLG \cite{bien-etal-2020-recipenlg} and include the survey of CHAMP \cite{mao2024champ} in Appendix~\ref{sec:CHAMP}. Finally, we conclude with a comprehensive ablation study of our proposed pipeline. For completeness, we also compare inference times for our system and the baselines in Appendix~\ref{app:runtime}.

\subsection{Baselines}
\label{sec:baselines}

We compare our system against widely used natural language generation approaches that involve prompting an LLM. For our experiments, we use \texttt{gpt-3.5-turbo-0125} from OpenAI as the language model for all approaches.
\begin{itemize}[leftmargin=*]\vspace{-1mm}
    \item\textbf{Zero-Shot}: We prompt the language model to generate the steps for a procedure, given the task instructions and the procedure's input and output.\vspace{-2mm}
    \item\textbf{Few-Shot}: We sample $K=3$ random procedures from the training set and include those in the prompt after the task instructions, which are the same as in the zero-shot case.\vspace{-2mm}
    \item\textbf{RAG}: We retrieve $k=3$ procedures based on the provided input and output and then use those three procedures in the prompt like the few-shot case.\vspace{-2mm}
    \item\textbf{ReAct}: We adapt the popular ReAct system~\cite{reAct} to our task by prompting the model with access to the procedural knowledge store as a ``tool'' which it can search, and allowing successive queries to the tool before deciding to produce a final answer.
\end{itemize}

\begin{table*}[]
\centering
\begin{tabular}{|c|ccc|ccc|}
\hline
\multirow{2}{*}{\begin{tabular}[c]{@{}c@{}}Dataset\\ Method\end{tabular}} & \multicolumn{3}{c|}{LCStep} & \multicolumn{3}{c|}{RecipeNLG}\\
\cline{2-7} & \multicolumn{1}{c|}{AAG win} & \multicolumn{1}{c|}{AAG loss} & Ties & \multicolumn{1}{c|}{AAG win} & \multicolumn{1}{c|}{AAG loss} & Ties \\ \hline
Zero-Shot                                                                    & \multicolumn{1}{c|}{\textbf{67.85}}   & \multicolumn{1}{c|}{26.78}             & 5.37          & \multicolumn{1}{c|}{\textbf{69.90}}   & \multicolumn{1}{c|}{18.35}             & 11.75         \\ \hline
Few Shot                                                                     & \multicolumn{1}{c|}{\textbf{71.42}}   & \multicolumn{1}{c|}{17.86}             & 10.72         & \multicolumn{1}{c|}{\textbf{81.70}}   & \multicolumn{1}{c|}{10.15}             & 8.15          \\ \hline
RAG                                                                          & \multicolumn{1}{c|}{\textbf{69.64}}   & \multicolumn{1}{c|}{12.50}             & 17.86         & \multicolumn{1}{c|}{\textbf{85.70}}   & \multicolumn{1}{c|}{7.60}              & 6.70          \\ \hline
ReAct                                                                        & \multicolumn{1}{c|}{\textbf{98.21}}   & \multicolumn{1}{c|}{1.79}              & 0             & \multicolumn{1}{c|}{\textbf{97.65}}   & \multicolumn{1}{c|}{1.55}              & 0.80          \\ \hline \hline
NO QG                                                                    & \multicolumn{1}{c|}{\textbf{71.43}}   & \multicolumn{1}{c|}{23.21}             & 5.36          & \multicolumn{1}{c|}{\textbf{80.05}}   & \multicolumn{1}{c|}{11.85}             & 8.10         \\ \hline
NO AG                                                                     & \multicolumn{1}{c|}{\textbf{39.28}}   & \multicolumn{1}{c|}{39.28}             & 21.44         & \multicolumn{1}{c|}{\textbf{58.45}}   & \multicolumn{1}{c|}{26.20}             & 15.35          \\ \hline
NO CR                                                                          & \multicolumn{1}{c|}{\textbf{69.64}}   & \multicolumn{1}{c|}{10.71}             & 19.65         & \multicolumn{1}{c|}{\textbf{87.15}}   & \multicolumn{1}{c|}{7.20}              & 5.65          \\ \hline
NO AG - NO CR                                                                        & \multicolumn{1}{c|}{\textbf{76.78}}   & \multicolumn{1}{c|}{16.07}              & 7.15             & \multicolumn{1}{c|}{\textbf{88.30}}   & \multicolumn{1}{c|}{6.45}              & 5.25          \\ \hline
\end{tabular}
\caption{Results of the LLM-based pairwise evaluation as discussed in Sec \ref{sec:eval-method}. \textit{AAG win} signifies that the LLM chose AAG over the other method, \textit{AAG loss} signifies the vice-versa, and \textit{Ties} implies that the language model was indifferent to the two options. The first four rows correspond to a comparison with the baselines, whereas the last four rows correspond to different ablations of AAG (see sec~\ref{sec:ablations}). The reported numbers are in percentage. AAG is preferred substantially more than all its counterparts, even on the CHAMP dataset, as presented in appendix~\ref{sec:CHAMP}.}
\label{tab:pair-eval}
\end{table*}

\subsection{Evaluation Methodology}
\label{sec:eval-method}
We compare the proposed AAG system and the other methods under an LLM-based pairwise evaluation scheme. We prompt the LLM to evaluate the two procedures (AAG and other method) on:
\begin{itemize}[leftmargin=*]\vspace{-2mm}
    \item Their ability to accomplish the specified goal;\vspace{-2mm}
    \item The level of details, clarity, and coherence of the steps;\vspace{-2mm}
    \item Coverage of the resources specified in user input.\vspace{-2mm}
\end{itemize}

The exact prompt for this evaluation can be found in the appendix \ref{sec:pair-eval}. Note that we don't provide ground-truth steps as input in the prompt due to several reasons: 1) there can be multiple correct procedures to achieve a single task, 2) the ground-truth procedures are seldom incomplete and noisy, specifically in recipes as they are scraped from the internet, and 3) to prevent biasing the LLM towards only matching the keywords in the generated and ground-truth procedure steps.

Both procedures under comparison are generated by the same LLM; thus, LLM's bias towards its own generation gets nullified. To further control for randomness and de-bias the LLM evaluation towards ordering procedures in the prompt, we run ten prompt requests for each sample, with five requests having the AAG procedure placed first and the other 5 have the other method placed first. Moreover, the LLM is set to a temperature of 0.7, and each of the ten calls is on a different seed of the LLM, but the ten seeds remain fixed during the evaluation of all the samples. Finally, we obtain the result for a sample using majority voting among the results of 10 LLM calls.

We must note that we report a pairwise comparison of methods and AAG in table \ref{tab:pair-eval}, \ref{tab:champ-res} and \ref{tab:n-query}; therefore, the numbers cannot be compared across methods. A higher number for a technique means that the LLM is more \textit{confident} that one system is better than the other in the pair, not that one system is better than the other systems not in the pair.

\subsection{System Design Choices}
\label{sec:design-choice}
The AAG system has no trainable parameters and relies on frozen LLMs. The temperature of the OpenAI LM is set to a value of 0.7 for all evaluations. We use the \textit{all-mpnet-base-v2} model from sentence-transformers \cite{reimers-2019-sentence-bert} as the embedding model for procedural memory Weaviate store, creating dense 768-dimensional embeddings for each procedure object. All three data fields of a procedure (input, output, and steps) are embedded together in one embedding.

\subsection{LCStep: An Unfamiliar Domain}
\noindent \textbf{Setup:} We sorted the 276 procedures in LCStep by increasing length and selected the final (longest) 56 examples as the test set, with the preceding 27 examples as a validation set. The remaining examples populate the procedural memory store. Selecting the longest examples for the test set evaluates AAG's ability to leverage simpler procedural knowledge to propose solutions for more complicated tasks.\\

\noindent\textbf{Results:} From Table~\ref{tab:pair-eval}, it is evident that the LLM prefers the procedures generated by the AAG system over all other baselines by a substantial margin (>40\%). Figure~\ref{fig:lcstep-qual-example} shows a qualitative example of how AAG manages to provide more of the detail present in the reference procedure. While the RAG approach stops at just initializing agents, the AAG output, similar to the ground truth, takes a step ahead and encourages the user to test the implemented setup for correctness, which is a good coding practice.

\subsection{RecipeNLG: A Familiar Domain}
\noindent \textbf{Setup:} Under our formalism, we construct a procedure using the recipe title as output, the ingredients as input, and the directions as the procedural steps. From over 2 million recipes, we select a random subset of 10000 recipes for our experiments (fixed across methods). Out of these 10000 recipes, we select 2000 as test and 1000 as validation examples. The remaining 7000 examples populate the procedural memory store.\\

\noindent \textbf{Results:} As highlighted in Table~\ref{tab:pair-eval}, the AAG system outperforms its counterpart for all methods on the RecipeNLG dataset. Interestingly, even though recipes are already familiar to the LLM, the analogical reasoning in the AAG pipeline further improves the quality of the generated responses according to our evaluations. Thus, while the AAG system is expected to shine better on specialized unseen domains like LCStep, it still provides clear benefits for a familiar domain.

Since cooking is more common knowledge than LangChain, we were able to conduct a blind human study to corroborate our pairwise evaluation results. We surveyed human raters to compare generations from AAG and RAG for 50 random test examples from the RecipeNLG test set. 3-6 human raters evaluated each pair of generations. The study revealed that users preferred AAG recipes in \textbf{56\%} of instances and preferred RAG recipes in only 16\% of examples. (28\% of examples resulted in a tie.) This underscores the conclusions of the pairwise evaluation on the efficacy of the AAG system on the RecipeNLG dataset. See Appendix~\ref{sec:qual-recipe} for a side-by-side comparison of generated outputs for a recipe from the test set.

\begin{figure*}[!ht]
\centering
    \begin{tabular}{|>{\columncolor{soft}}p{0.3\linewidth}|>{\columncolor{soft}}p{0.3\linewidth}|>{\columncolor{soft}}p{0.3\linewidth}|}
    \hline
    \multicolumn{3}{|>{\columncolor{highlight}}l|}{\textbf{Output:} set up a custom input schema for a tool with strict requirements and custom validation logic} \\
    \multicolumn{3}{|>{\columncolor{highlight}}l|}{\textbf{Input:} an LLM} \\ \hline \vspace{1mm}
    {\small \textbf{Reference}}\vspace{2mm}
    {\tiny
    \begin{itemize}[leftmargin=*, itemsep=2pt, parsep=2pt, topsep=2pt, partopsep=2pt]
        \item Define a class \texttt{ToolInputSchema} that inherits from \texttt{pydantic.BaseModel}. Include the fields you require, in this case a URL, and a root validator method that checks the domain of the URL against a list of approved domains.
        \item Initialize your tool (e.g.~\texttt{langchain.tools.requests. tool.RequestsGetTool}) with \texttt{ToolInputSchema} as args\_schema and requests wrapper (e.g.~\texttt{langchain.tools.requests.tool. TextRequestsWrapper}).
        \item Initialize your agent with \texttt{langchain.agents. initialize\_agent}, providing the tool, the language model (e.g.~\texttt{langchain.llms.OpenAI}), and the agent type (e.g.~\texttt{langchain.agents.AgentType. ZERO\_SHOT\_REACT\_DESCRIPTION}).
        \item Run the agent with a valid request using the `run` method. For example, a request to get the main title of an approved website.
        \item If you try to run the agent with a request for a non-approved website, it will raise a \texttt{ValidationError}.
    \end{itemize}
    }
    & \vspace{1mm}
    {\small \textbf{RAG}}\vspace{2mm}
    {\tiny
    \begin{itemize}[leftmargin=*, itemsep=2pt, parsep=2pt, topsep=2pt, partopsep=2pt]
        \item Define a function for your tool that takes the custom input schema and includes custom validation logic.
        \item Create a \texttt{StructuredTool} using \texttt{langchain.tools. StructuredTool.from\_function}, providing the function you defined in step 1.
        \item Initialize your agent with \texttt{langchain.agents. initialize\_agent}, providing a list containing the \texttt{StructuredTool}, the language model, and the agent type \texttt{langchain.agents.AgentType. STRUCTURED\_CHAT\_ZERO\_SHOT\_REACT\_DESCRIPTION}.
    \end{itemize}
    } & \vspace{1mm}
    {\small \textbf{AAG}}\vspace{2mm}
    {\tiny
    \begin{itemize}[leftmargin=*, itemsep=2pt, parsep=2pt, topsep=2pt, partopsep=2pt]
        \item Define a function for the tool with custom validation logic to enforce strict requirements, e.g.~a function \texttt{custom\_input\_function(input1, input2)} with specific validation conditions.
        \item Create a \texttt{StructuredTool} using the defined function with \texttt{langchain.tools.StructuredTool. from\_function}.
        \item Define a custom LLM class inheriting from \texttt{langchain. llms. base. LLM} and implement \texttt{\_llm\_type}, \texttt{\_call}, and \texttt{\_identifying\_params} methods.
        \item Instantiate the custom LLM class with any necessary parameters like validation rules.
        \item Initialize the agent with \texttt{langchain.agents. initialize\_agent} providing the custom LLM, the language model, and the agent type \texttt{langchain.agents.AgentType. STRUCTURED\_CHAT\_ZERO\_SHOT\_REACT\_DESCRIPTION}.
        \item Test the setup by running sample inputs through the tool to confirm the custom input schema and validation logic function correctly.
    \end{itemize}
    } \\ \hline
    \end{tabular}
  \caption {Qualitative example comparing the ground-truth steps and the steps generated by RAG and the AAG system on the LCStep dataset. The AAG procedure includes more details like function names and signatures and good practices like testing the built systems, in contrast to the RAG system.}\vspace{-3mm}
  \label{fig:lcstep-qual-example}
\end{figure*}

\subsection{Ablation Study}
\label{sec:ablations}
We perform a rigorous ablation study of our proposed AAG architecture. We experiment with removing the three architectural contributions in all permutations, i.e., the query generator, answer generator, and critic-guided iterative refinement. We describe each of the resulting modifications below:
\begin{itemize}[leftmargin=*]\vspace{-1mm}
    \item \textbf{AAG-NOQG}: This represents the AAG system without the query generator. The absence of a query generator also implies the absence of the pipeline's answer generator and Update Response modules. The RAG response is directly fed into critic-based iterative refinement, to produce the final output.\vspace{-2mm}
    \item \textbf{AAG-NOAG}: This represents the AAG system without the answer generator module. For each question output by the query generator, we concatenate all the retrieved procedures to form the answer, removing the duplicates across queries and forming one big context with all the procedures.\vspace{-2mm}
    \item \textbf{AAG-NOCR}: This represents the AAG system without critic-guided iterative refinement. The Update Response module's output is considered the system's final output.
    \item \vspace{-2mm} \textbf{AAG-NOAG-NOCR}: This represents the AAG system without the answer generator and the critic-guided iterative refinement. The Update Response module receives the context as in AAG-NOAG; its output is the system's final output.
\end{itemize}

\noindent We also vary the number of queries $N$ generated by the query generator and include its observations in appendix \ref{sec:n-query-ablation}. \\

\noindent \textbf{Results}: The lower half of Table~\ref{tab:pair-eval} displays the results for the pairwise comparison between AAG and each of the ablation methods above on both datasets. From the table, it can be observed that AAG performs better than any of its ablation variants, highlighting the importance of each component in the pipeline. The numbers for the AAG-NOQG variant showcase the analogy's significance in our system, achieved by the query and answer generator modules.
In the absence of any of the three modules, namely the query, answer generator, and iterative refinement, performance is reduced.

\section{Related Work}
Based on the \textit{Rewrite-Retrieve-Read} framework \cite{ma2023query}, AAG differs from the conventional RAG \cite{lewis2021retrievalaugmented} in three key aspects: sophisticated prompting approach, analogical reasoning, and the iterative workflow for response generation (Section \ref{sec:overall-workflow}), in addition to the application in knowledge-intensive procedural QA. We refer the readers to \cite{gao2024retrievalaugmented} for a survey covering the latest RAG research.\\

\noindent \textbf{Standard Prompting Methods}: Zero-shot prompting~\cite{chang2024efficient}, few-shot prompting~\cite{llms-are-few-shot-learners}, and RAG~\cite{lewis2021retrievalaugmented} are the most common LLM-based approaches to natural language generation. Few-shot prompting requires a fixed set of labeled examples, and RAG requires some supporting set of documents, not necessarily of the same form as the system's generated text. For example, when augmenting LLMs with general-knowledge Wikipedia articles to answer factual questions. In this work, we perform a sophisticated interleaving of prompts and memory searches to retrieve finer context information from the analogical examples in procedural memory.\\

\noindent \textbf{Analogical Reasoning in LLMs}: Many papers attempt to elicit reasoning in LLMs by various prompting methods \cite{zhou2024selfdiscover,wang-etal-2023-chatgpt-defend}. Specific to analogical reasoning, early attempts show by case studies that LLMs can do analogical reasoning just like humans \cite{webb2023emergent}. Recent works explore analogy generation and reasoning with knowledge graphs on LLMs \cite{yuan2023analogykb, bhavya2023cam, bhavya2022analogy}. These works are applied to propositional knowledge and rely on large-scale external knowledge bases to store entity relationships and perform analogical reasoning. Yu et al.~(\citeyear{thoughtPropagation}) and Yasunaga et al.~(\citeyear{yasunaga2023large}) prompt the model to propose problems (and their solutions) analogous to the original problem, to inform the final solution. Unfortunately this approach is prone to failure on under-represented or unseen domains due to a lack of knowledge. Thus, in our work, we cast the problem as a domain adaptation problem where we retrieve real solutions to analogous real problems from the procedural memory, thus receiving performance benefits in both familiar and unfamiliar knowledge-intensive domains.\\

\noindent \textbf{Iterative Pipeline}: Numerous recent works based on the traditional \textit{Retrieve-then-Read} paradigm undertake an iterative workflow to generate the response to an input query. Asai et al.~(\citeyear{asai2023selfrag}) train a separate LM to predict reflection tokens like `retrieve,' `relevant,' and `irrelevant' within the response, which determines the subsequent control flow to generate or refine subsequent chunks of the response accordingly. On the contrary, Shao et al.~(\citeyear{shao-etal-2023-enhancing}) intersperse Generation-Augmented Retrieval (GAR) and RAG to sequentially expand the search query in the memory and obtain relevant information to plug into the final response. Similar to Shao et al.~(\citeyear{shao-etal-2023-enhancing}), we expand the search query to meaningful sub-questions that help retrieve relevant knowledge from the procedural memory at the granularity of individual steps.

\section{Conclusion}

In this work, we introduced a novel formalism for procedural knowledge inspired by human reasoning and extended it to build a novel LLM-based system to perform analogical reasoning for procedural Q\&A. Our experiments demonstrate its increased performance over standard RAG on three domains. The system performed especially well on our self-curated LCStep dataset, showing that AAG can better augment a frozen LLM with the knowledge needed to answer questions on unseen domains. Although recipes are highly familiar to general-purpose LLMs like GPT-3.5, the AAG system still significantly improves the model outputs even according to human review. These results show that the structured procedural memory, analogical reasoning through the combination of query and answer generator, and iterative critique/refinement all bolster the performance of our system on procedure generation tasks. Our theoretical framework provides a basis for solving high-level planning tasks more effectively in varied domains.

\section*{Limitations and Future Work}
\label{sec:limitations}

This framework deliberately ignores procedure structure that is non-linear, meaning that a non-linear procedure must express its non-linearity in the text of the steps. This is a reasonable assumption for many tasks performed by humans, but not for computer algorithms. In future work we will expand our procedure definition to allow dependencies to be represented so that analogy can be done on the procedure structure beyond just step ordering. In particular, we plan to more tightly couple our system with the formal structure we have proposed, by leveraging the composable nature of our procedure definition to perform finer-grained search and reasoning over sub-procedures. This may involve a sort of subroutine representation similar to the skill library from Voyager~\cite{wang2023voyager}.

As mentioned in Section~\ref{sec:lcstep}, the window is limited for using LCStep to measure system performance on unseen domains. In this work it was necessary for us to use an older LLM in order to test this for AAG. Future work can add another incremental LCStep dataset with updated instructions for a later version of LangChain, which would allow us to test how systems perform as documentation is updated.
\section*{Ethical Considerations}
\label{sec:ethics}

The use of automated systems to generate and execute plans can and will result in unintended consequences, whether humans are in the loop or on the loop~\cite{LeinsKaspersen2021}. In addition, automating actions generally done by humans (here planning and reasoning) has the social effect of anthropomorphizing these systems, leading to a gradual shift in accountability from human stakeholders onto the systems themselves~\cite{weidinger2021ethical}.

The domains which we have applied our system to in this work do not constitute immediate risks. For LCStep, the generated steps cannot easily be converted into code, although this could become increasingly easy as planning-based code generation continues to improve~\cite{code-planning}. Mathematical problem solving poses no real-world risks.

Thanks to Google's AI Overviews feature, there has already been at least one instance of a generative AI system producing a food recipe that would be harmful if followed~\cite{glue-in-pizza}.\footnote{See \url{https://web.archive.org/web/20240222195427/https://stayhealthyhacks.com/can-you-freeze-off-a-mole/} for another humorous example of incorrect food-related content generated by AI.} While it might be obvious to most that it's a bad idea to add glue to food, it's clear that automated recipe generation could result in worse than a botched meal. No matter how many safeguards are added, systems that generate content ought to appropriately inform users of the risks of relying on the content without verification. We hope that the general public continues to be made aware of humorous mistakes like this so that people learn to approach AI content with healthy skepticism.

\nocite{pmlr-v202-shrivastava23a}
\bibliography{custom}
\newpage
\clearpage
\appendix

\section*{Appendix}

\section{LCStep dataset creation details}
\label{app:lcstep}
LCStep contains three sets of documents: API reference, conceptual documentation, and procedures. In this paper, we only used the procedures, but we plan to leverage the supporting material in future systems. See Figure~\ref{fig:dataset-workflow} for a diagram of the process of generating the LCStep data.

As Langchain was unstable and changing quickly at the time we collected the data, we made sure to capture the state of the code and documentation at \href{https://github.com/langchain-ai/langchain/releases/tag/v0.0.249}{version 0.0.249}.

\begin{figure*}
   \centering
   \includegraphics[width=\linewidth]{dataset\_workflow.drawio.png}
   \caption{The workflow used to generate the LCStep dataset.}\label{fig:dataset-workflow}
\end{figure*}

\subsection{API Reference}

We generate the API reference material from the source files in the LangChain \href{https://github.com/langchain-ai/langchain/}{GitHub repository} using Sphinx. These files contain descriptions of all APIs in the Python package, including call signatures and argument descriptions. These files do not contain any usage examples or high-level explanation.

\subsection{Conceptual Documentation and Procedures}

We collected these resources by scraping the \href{https://python.langchain.com/}{Langchain Python docs} when v0.0.249 was the version live on the website. We manually filter out topic pages and stubs, leaving 228 documents. We then manually classified these into around 30 documents of conceptual documentation, and around 180 documents containing tutorials/guides.

For the 180 tutorials/guides, we prompted GPT-4 (see Listing~\ref{lst:lcstep-formatting}) to extract a list of high-level steps necessary to accomplish the goal. We then prompted GPT-4 (see Listing~\ref{lst:lcstep-check}) to rate those extracted procedures using a list of criteria. We found that this caught many mistakes where GPT-4 did not follow all the stated instructions. In those cases, we had the model revise the steps to meet the requirements, and then we manually checked the revised versions.

\begin{lstlisting}[caption={prompt for extracting procedures},label={lst:lcstep-formatting},frame=single,breaklines=true,breakindent=0pt,basicstyle=\small\ttfamily]
    You are helping convert tutorials for a Python library into step-by-step procedures. The tutorials describe how to perform various tasks with code and examples that use relevant APIs in the library. Your task is to simplify a tutorial into two parts: a description of the specific goal, and a minimal set of the necessary instructions. Write "Goal: {description}", and use ordered numbering for the steps.

    The goal should be a complete and specific task rather than a general approach, e.g. "integrate a calculator into a chat model" instead of "integrate custom functions with OpenAI functions agent". You usually should not refer to specific APIs in the goal description; focus on describing functionality rather than a particular approach.

    Each instruction is a logically distinct action, and explicitly refers to the library methods used for that action. Remove all unnecessary details, code, parameters, and example usage; keep only the information strictly necessary to accomplish the procedure. A side note with auxiliary info may be provided after the steps, but this note must not contain information relevant to accomplishing the described goal.

    Do not mention importing the required modules as a separate step. Instead, always use the full path to APIs (e.g. `langchain.chat_models.ChatOpenAI`) when referring to them in the steps. Do not mention specific API parameters unless they're important to understanding how the procedure fits together, e.g. don't include things like the verbosity or the model temperature unless those matter for accomplishing the stated goal.

    If the raw text actually contains more than one logically distinct procedure, you may then output multiple sets of goals and instructions for each different procedure (delimited by NEW PROCEDURE), but this should only be done if the goals are fundamentally different. Don't create separate procedures if one procedure could be expressed in a side note. These separate procedures must not make reference to one another, as in "follow steps 2-4 from the previous procedure".
\end{lstlisting}

\begin{lstlisting}[caption={prompt for checking procedures},label={lst:lcstep-check},frame=single,breaklines=true,breakindent=0pt,basicstyle=\small\ttfamily]
You are judging the output of another language model on a particular task. The original instructions are provided below:

BEGIN INSTRUCTIONS
{instructions from Listing 1}
END INSTRUCTIONS

Below is an example of how you should judge the text. Please list whether the provided text passes or fails each requirement in the list:

- [PASS] format is as expected
- [PASS] the goal is a specific task rather than general approach
- [PASS] the resources line includes only the required external APIs or models that must be accessible for the steps to work
- [PASS] the steps fully achieve the task described in the goal
- [PASS] does not unnecessarily refer to APIs in the goal description
- [PASS] instructions are logically distinct
- [FAIL] no unnecessary details, code, or parameters like verbosity or temperature, etc. (INSERT REASON FOR FAIL)
- [PASS] side note is relevant if present
- [PASS] all API references use the full path
- [PASS] if there are multiple procedures, they're describing two independent processes, not two sub processes that succeed one another
- [PASS] if there are multiple procedures, they're describing two different goals, not the same process with a simple variation

If any requirement fails, correct the procedure by rewriting it in a manner that passes all checks. If either of the last two checks fail, join the procedures together and specify any necessary variation in the instructions or a side note.
\end{lstlisting}

\section{CHAMP: Challenging Math Problems Dataset}
\label{sec:CHAMP}
\noindent \textbf{Setup:} CHAMP is a dataset of 270 competition-level math problems annotated with concepts, hints, and step-by-step solutions \cite{mao2024champ}. For our task definition, we use the problem statement as the output, the hints and problem category as input, and the step-by-step solutions as the procedure steps. We also append ``The answer is \textit{<answer>}'' as a final step in the procedure. We randomly shuffle the problems and choose 54 test examples and 27 validation examples. Random shuffling is important to ensure diversity so that each data split contains problems of all types. The remaining problems populate the procedural memory store.\\

\begin{table}[]
\centering
\begin{tabular}{|c|ccc|}
\hline
\multirow{2}{*}{\begin{tabular}[c]{@{}c@{}}Dataset\\ Method\end{tabular}} & \multicolumn{3}{c|}{CHAMP}\\ \cline{2-4}  & \multicolumn{1}{c|}{AAG win}        & \multicolumn{1}{c|}{AAG loss}       & Ties  \\ \hline
Zero-Shot                                                                    & \multicolumn{1}{c|}{\textbf{55.55}} & \multicolumn{1}{c|}{37.03}          & 7.42  \\ \hline
Few Shot                                                                     & \multicolumn{1}{c|}{42.59}          & \multicolumn{1}{c|}{\textbf{46.29}} & 11.12 \\ \hline
RAG                                                                          & \multicolumn{1}{c|}{\textbf{53.70}} & \multicolumn{1}{c|}{31.48}          & 14.82 \\ \hline
ReAct                                                                        & \multicolumn{1}{c|}{\textbf{68.51}} & \multicolumn{1}{c|}{25.92}          & 5.55  \\ \hline \hline
NO QG                                                                    & \multicolumn{1}{c|}{\textbf{51.85}} & \multicolumn{1}{c|}{37.03}          & 11.12  \\ \hline
NO AG                                                                     & \multicolumn{1}{c|}{33.33}          & \multicolumn{1}{c|}{\textbf{55.55}} & 11.12 \\ \hline
NO CR                                                                          & \multicolumn{1}{c|}{\textbf{48.15}} & \multicolumn{1}{c|}{38.88}          & 12.97 \\ \hline
NO AG - NO CR                                                                        & \multicolumn{1}{c|}{\textbf{50.00}} & \multicolumn{1}{c|}{37.03}          & 12.97  \\ \hline

\end{tabular}
\caption{Results of the LLM-based pairwise evaluation as discussed in Sec \ref{sec:eval-method}. As the name suggests, \textit{AAG win} signifies that the LLM chose AAG over the other method, \textit{AAG loss} signifies the vice-versa, and \textit{Ties} implies that the language model was indifferent to the two options. The first four rows correspond to a comparison with the baselines, whereas the last four rows correspond to different ablations of AAG (see sec \ref{sec:ablations}). The reported numbers are in percentage. LLM prefers AAG responses more than most of its counterparts.}
\label{tab:champ-res}
\end{table}

\noindent \textbf{Results:} The results in table \ref{tab:champ-res} suggests that the LLM prefers the AAG responses more than most of its counterparts under the pairwise evaluation scheme. The most interesting observation from table \ref{tab:champ-res} is that the AAG system performs slightly worse than the few-shot system. Aligned with this observation is another observation in the ablations, where removing answer generator module improves the performance of the system. This throws light on the fact that steps in maths solutions share far more context than the other domains, and hence, any attempt of filtering out content, takes away some crucial information, thus harming the performance. Despite this, AAG still manages to outperform traditional RAG, showcasing the effectiveness of multiple analogical retrievals from the memory.

\section{Qualitative Example for RecipeNLG}
\label{sec:qual-recipe}
\begin{figure*}[!ht]
\centering
    \begin{tabular}{|>{\columncolor{soft}}p{0.3\linewidth}|>{\columncolor{soft}}p{0.3\linewidth}|>{\columncolor{soft}}p{0.3\linewidth}|}
    \hline
    \multicolumn{3}{|>{\columncolor{highlight}}l|}{\textbf{Output:} Autumn Soup(Microwave Recipe)} \\
    \multicolumn{3}{|>{\columncolor{highlight}}p{0.96\linewidth}|}{\textbf{Input:} 1 lb. lean ground beef, 1 c. chopped onion, 1 c. chopped celery, 4 c. hot water, 2 c. 1/2-inch potato cubes, 1 c. thinly sliced carrots, 1 tsp. salt, 1/2 tsp. dried basil leaves, 1/4 tsp. ginger, 1 bay leaf, 3 tomatoes, cut into eighths and sliced in half} \\ \hline \vspace{1mm}
    {\small \textbf{Reference}}\vspace{2mm}
    {\tiny
    \begin{itemize}[leftmargin=*, itemsep=2pt, parsep=2pt, topsep=2pt, partopsep=2pt]
        \item Mix ground beef, onion and celery in 5-quart casserole dish.
        \item Microwave at High (100\%) until ground beef loses its pink color, 5 to 7 minutes, stirring after half of cooking time.
    \end{itemize}
    }
    & \vspace{1mm}
    {\small \textbf{RAG}}\vspace{2mm}
    {\tiny
    \begin{itemize}[leftmargin=*, itemsep=2pt, parsep=2pt, topsep=2pt, partopsep=2pt]
        \item In a microwave-safe dish, combine lean ground beef, chopped onion, and chopped celery. Microwave on high (at 70\% power) for about 5-7 minutes or until the beef is browned and vegetables are tender.
        \item Add hot water, potato cubes, sliced carrots, salt, dried basil leaves, ginger, and bay leaf to the dish. Microwave on high (at 80\% power) for approximately 10-12 minutes, stirring occasionally, or until the potatoes are tender.
        \item Add tomatoes to the soup and microwave on high (at 70\% power) for about 3-5 minutes or until heated through.
        \item Adjust seasoning to taste. Let the soup rest for a few minutes before serving to allow flavors to meld together.
        \item Serve the autumn soup hot.
    \end{itemize}
    } & \vspace{1mm}
    {\small \textbf{AAG}}\vspace{2mm}
    {\tiny
    \begin{itemize}[leftmargin=*, itemsep=2pt, parsep=2pt, topsep=2pt, partopsep=2pt]
        \item In a large microwave-safe casserole dish of at least 2-quart capacity, combine the lean ground beef, chopped onion, and chopped celery. Microwave on high until the beef is cooked through, breaking it up as it cooks. Look for the beef to be browned and no longer pink.
        \item Add hot water, potato cubes, sliced carrots, salt, dried basil leaves, ginger, and bay leaf to the casserole dish. Stir well to combine all the ingredients thoroughly.
        \item Cover the casserole dish with a microwave-safe lid or microwave-safe plastic wrap, leaving a small vent for steam to escape. Microwave on high for about 15 minutes or until the vegetables are tender. Check the tenderness of the vegetables with a fork.
        \item Add the tomatoes to the soup, stir, and microwave for an additional 2-3 minutes until the tomatoes are heated through.
        \item Remove the bay leaf before serving. Taste the soup and adjust seasoning with salt and pepper if needed. Enjoy your Autumn Soup straight from the microwave.
    \end{itemize}
    } \\ \hline
    \end{tabular}
  \caption {Qualitative example comparing the reference steps with the steps generated by RAG, and the AAG system on an example from the RecipeNLG dataset. The AAG procedure includes more details on how to perform each step and is easier to follow, for example, checking the tenderness of vegetables and how to serve them. The baselines, on the other hand, suggest steps but on a coarser granularity.}\vspace{-3mm}
  \label{fig:recipe-qual-example}
\end{figure*}

Figure~\ref{fig:recipe-qual-example} shows the reference steps to prepare a meal named ``Autumn Soup'' and the steps generated by RAG and AAG. Many recipes in the dataset are relatively short, and LLMs tend to overproduce text in comparison. RAG and AAG tend to include more details about each step, thus ensuring that a user knows how to execute the step without any more questions or confusion. AAG, especially, provides more holistic information, like the ways to check tenderness of the vegetables after they have been cooked, which is left out by the RAG system.

\section{Retrieved procedures for Qualitative Examples}
\label{sec:fig-details}
Figures \ref{fig:lcstep-qual-example} and \ref{fig:recipe-qual-example} emphasize the high-quality generations by AAG system as compared to the baselines. Distinctively, AAG includes a lot of additional details and information about the procedure steps. To motivate that the source of this additional knowledge is the retrieved procedures from memory and not LLM hallucinations, below is the list of queries generated by query generator and correspondingly retrieved procedures for both the examples:\\

\noindent \textbf{Figure 3}
\begin{itemize}[leftmargin=*]
    \item Microwave cooking times for ground beef
    \begin{itemize}
        \item Ground Beef Casserole OAMC
        \item Ground Beef Casserole
        \item Ground Beef Casserole
    \end{itemize}
    \item How to properly drain excess fat from cooked ground beef?
    \begin{itemize}
        \item Ground Beef Casserole OAMC
        \item Ground Beef Casserole
        \item Ground Beef Casserole
    \end{itemize}
    \item Microwave cooking times for potatoes and carrots
    \begin{itemize}
        \item Braised Potatoes And Carrots
        \item Microwave Scalloped Potatoes
        \item Fast Roasted Potatoes With Paprika
    \end{itemize}
    \item How to know when vegetables are tender in the microwave?
    \begin{itemize}
        \item Microwave Scalloped Potatoes
        \item Fast Roasted Potatoes With Paprika
        \item Braised Potatoes And Carrots
    \end{itemize}
\end{itemize}

\noindent (Note that some titles are the same, but they are different recipes with different sets of ingredients.)\\

\noindent \textbf{Figure 5:}
\begin{itemize}[leftmargin=*]
    \item How to define custom input schema requirements?
    \begin{itemize}
        \item set up a tool that requires multiple inputs for an agent
        \item set up a tool that requires a single string input for an agent
        \item use a Pydantic parser to query a language model for JSON outputs that conform to a specific schema
    \end{itemize}
    \item Methods for designing custom validation logic for input schemas.
    \begin{itemize}
        \item refine the tagging of a passage by specifying possible values, descriptions, and required properties for each attribute in the schema
        \item create a system to interact with a SQL database using natural language queries, with a custom prompt and query checker
        \item solve a 4x4 Sudoku puzzle using a Tree of Thought (ToT) guided by a Large Language Model
    \end{itemize}
    \item How to implement a custom input schema using an LLM?
    \begin{itemize}
        \item set up a tool that requires multiple inputs for an agent
        \item create a custom LLM that returns the first N characters of the input
        \item Create a custom MRKL agent with a custom LLMChain that takes multiple inputs, capable of answering queries in a specific language
    \end{itemize}

    \item Best practices for testing custom input schemas.
    \begin{itemize}
        \item test an agent using a fake Language Learning Model (LLM) that simulates responses
        \item Simulate human responses for testing a specific feature of a software using a pseudo LLM
        \item create a system to interact with a SQL database using natural language queries, with a custom prompt and query checker
    \end{itemize}
\end{itemize}

\section{Ablation for Number of Re-written Queries}
\label{sec:n-query-ablation}
\begin{table*}[]
\centering
\begin{tabular}{|c|ccc|ccc|ccc|}
\hline
\multirow{2}{*}{\begin{tabular}[c]{@{}c@{}}Dataset\\ Method\end{tabular}} & \multicolumn{3}{c|}{LCStep}                                                 & \multicolumn{3}{c|}{RecipeNLG}                                                  & \multicolumn{3}{c|}{CHAMP}\\
\cline{2-10} & \multicolumn{1}{c|}{AAG win}        & \multicolumn{1}{c|}{AAG loss} & Ties  & \multicolumn{1}{c|}{AAG win}       & \multicolumn{1}{c|}{AAG loss}       & Ties & \multicolumn{1}{c|}{AAG win} & \multicolumn{1}{c|}{AAG loss}       & Ties  \\ \hline
N  = 3                                                                       & \multicolumn{1}{c|}{\textbf{50.00}} & \multicolumn{1}{c|}{28.57}    & 21.43 & \multicolumn{1}{c|}{38.60}         & \multicolumn{1}{c|}{\textbf{39.80}} & 21.6 & \multicolumn{1}{c|}{35.18}   & \multicolumn{1}{c|}{\textbf{50.00}} & 14.82 \\ \hline
N = 6                                                                        & \multicolumn{1}{c|}{\textbf{48.21}} & \multicolumn{1}{c|}{33.93}    & 17.86 & \multicolumn{1}{c|}{\textbf{43.4}} & \multicolumn{1}{c|}{37.8}           & 18.8 & \multicolumn{1}{c|}{35.18}   & \multicolumn{1}{c|}{\textbf{53.70}} & 11.11 \\ \hline
\end{tabular}
\caption{Comparison of the proposed AAG system with $N = 4$ queries generated by the query generator against the AAG system with $N = 3, 6$ queries respectively under the pairwise evaluation scheme (see sec \ref{sec:eval-method}). As the name suggests, \textit{AAG win} signifies that the LLM chose AAG over the other method, \textit{AAG loss} signifies the vice-versa, and \textit{Ties} implies that the language model was indifferent to the two options. The reported numbers are in percentage. LLM prefers AAG responses more than most of its counterparts on RecipeNLG and LCStep datasets.}
\label{tab:n-query}
\end{table*}

In this study, we aim to investigate the performance of the AAG system on changing the hyper-parameter for the number of queries $N$ generated by the query generator module. Traditionally, AAG uses 4 queries but we experiment with 3 and 6 queries in this evaluation. Table \ref{tab:n-query} shows the results for this ablation under the pairwise LLM evaluation. We note from the table that while 4 queries is preferred than 3 or 6 for RecipeNLG and LCStep, CHAMP narrates a mixed story with both 3 and 6 queries performing better. We attribute this mixed result to the high complexity of the dataset overall and the general-purpose LLM's limitation to rephrase the question into meaningful queries. Thus, it can be safely concluded that this is not a very sensitive hyper-parameter for our proposed system and setting a value of 4 only increases the performance as compared to its counterparts.

\section{Inference Time Comparison}
\label{app:runtime}
\begin{table*}[]
\centering
\begin{tabular}{|c|c|c|c|}
\hline
Method, Dataset & LCStep & RecipeNLG & CHAMP  \\ \hline
RAG             & 2.94s  & 2.78s     & 2.26s  \\ \hline
ReAct           & 2.40s  & 2.39s     & 4.18s  \\ \hline
AAG             & 23.44s & 22.53s    & 24.27s \\ \hline
\end{tabular}
\caption{Average inference time per example (in seconds) of the different systems on all the three datasets}
\label{tab:runtime}
\end{table*}

Table \ref{tab:runtime} shows the average inference time of the RAG, ReAct, and AAG systems on a sample from each of the three datasets. It can be observed that AAG takes roughly 7x more time than the RAG system. AAG performs five instances of retrieval for a single query, compared to 1 memory search in RAG. Furthermore, AAG makes 13 LLM calls per query, in contrast to only 1 for RAG. During generation, ReAct dynamically decides whether to do a fresh memory search or continue generating. Hence, the number of LLM calls and memory searches can vary across queries.

\section{Full Example of AAG with Prompts}
\label{sec:prompts}
Here we'll take an example from the LCStep test set. See the attached code for the prompts adapted to the other datasets. The output is \texttt{set up a custom input schema for a tool with strict requirements and custom validation logic}, and the input is \texttt{an LLM}. The reference steps are the following:

\begin{lstlisting}[caption={reference steps for this example},label={lst:aag-ref},frame=single,breaklines=true,breakindent=0pt,basicstyle=\small\ttfamily]
    1. Define a class `ToolInputSchema` that inherits from `pydantic.BaseModel`. Include the fields you require, in this case a URL, and a root validator method that checks the domain of the URL against a list of approved domains.
    2. Initialize your tool (e.g. `langchain.tools.requests.tool.RequestsGetTool`) with the `ToolInputSchema` as the args_schema and a requests wrapper (e.g. `langchain.tools.requests.tool.TextRequestsWrapper`).
    3. Initialize your agent with `langchain.agents.initialize_agent`, providing the tool, the language model (e.g. `langchain.llms.OpenAI`), and the agent type (e.g. `langchain.agents.AgentType.ZERO_SHOT_REACT_DESCRIPTION`).
    4. Run the agent with a valid request using the `run` method on your agent. For example, a request to get the main title of an approved website.
    5. If you try to run the agent with a request for a non-approved website, it will raise a `ValidationError`.
\end{lstlisting}

\noindent Given the input and output strings, the first prompt generates the queries:

\begin{lstlisting}[caption={query rewriting prompt},label={lst:aag-qr},frame=single,breaklines=true,breakindent=0pt,basicstyle=\small\ttfamily]
Please output high-level steps to complete the task below.

Then, given this high-level solution, think carefully step by step and provide 4 search engine queries for knowledge that you need to refine the solution to the question.

The output should be 'steps:' followed by a bulleted list with elements starting with '- ', and then 'queries:' followed by another bulleted list.

---

I want to create set up a custom input schema for a tool with strict requirements and custom validation logic using these resources: an LLM. 
\end{lstlisting}

\noindent The output is the following:

\begin{lstlisting}[caption={query rewriting output},label={lst:aag-qr-out},frame=single,breaklines=true,breakindent=0pt,basicstyle=\small\ttfamily]
  steps:
  - Understand the requirements and constraints for the tool's input schema.
  - Identify the necessary fields and validation rules based on the LLM.
  - Implement the custom input schema with the strict requirements and validation logic.
  - Test the custom input schema to ensure it functions correctly.

  queries:
  - How to define custom input schema in [tool name]?
  - Best practices for implementing custom validation logic in [tool name]?
  - How to integrate an LLM into custom input schema?
  - Common pitfalls to avoid when setting up a custom input schema in [tool name]?
\end{lstlisting}

\noindent Each of those queries is then used to retrieve 3 procedures from the procedure store. Those are listed here:

\begin{lstlisting}[caption={retrieved procedure titles for each query},label={lst:aag-ret},frame=single,breaklines=true,breakindent=0pt,basicstyle=\small\ttfamily]
- How to define custom input schema in [tool name]?
  - set up a tool that requires multiple inputs for an agent
  - build a structured tool chat agent capable of using multi-input tools and handling memory
  - set up a tool that requires a single string input for an agent
- Best practices for implementing custom validation logic in [tool name]?
  - create custom tools for a LangChain agent and handle tool errors
  - create a prompt template with variable validation
  - set up a tool that requires multiple inputs for an agent
- How to integrate an LLM into custom input schema?
  - set up a tool that requires multiple inputs for an agent
  - create a custom LLM that returns the first N characters of the input
  - Create a custom MRKL agent with a custom LLMChain that takes multiple inputs, capable of answering queries in a specific language
- Common pitfalls to avoid when setting up a custom input schema in [tool name]?
  - build a structured tool chat agent capable of using multi-input tools and handling memory
  - set up a tool that requires multiple inputs for an agent
  - create custom tools for a LangChain agent and handle tool errors
\end{lstlisting}

\noindent Then another prompt summarizes the knowledge. For the first generated query (\texttt{How to define custom input schema in [tool name]?}) that looked like this:

\begin{lstlisting}[caption={summarization prompt},label={lst:aag-su},frame=single,breaklines=true,breakindent=0pt,basicstyle=\small\ttfamily]
[Instruction]
You are a human expert whose job is to summarise the retrieved information below to answer the question. Please include the information only from the provided knowledge and make sure that the summary is complete, short and concise. Avoid introductory and closing lines at the start and end of your response. Don't directly refer to the titles in the provided knowledge when generating the summary.

---

[BEGIN QUESTION]
How to define custom input schema in [tool name]?
[END QUESTION]

[BEGIN INFORMATION]
DOCUMENTATION 'set up a tool that requires multiple inputs for an agent' using an LLM:

1. Define a function for your tool that takes multiple inputs.
2. Create a `StructuredTool` using `langchain.tools.StructuredTool.from_function`, providing the function you defined.
3. Initialize your agent with `langchain.agents.initialize_agent`, providing a list containing the `StructuredTool`, the language model, and the agent type `langchain.agents.AgentType.STRUCTURED_CHAT_ZERO_SHOT_REACT_DESCRIPTION`.

DOCUMENTATION 'build a structured tool chat agent capable of using multi-input tools and handling memory' using an LLM, PlayWrightBrowserToolkit:

1. Initialize a PlayWrightBrowserToolkit using `langchain.agents.agent_toolkits.PlayWrightBrowserToolkit.from_browser` with an asynchronous browser created using `langchain.tools.playwright.utils.create_async_playwright_browser`.
2. Get the tools from the browser toolkit using the `get_tools` method.
3. Initialize your language model using `langchain.chat_models.ChatOpenAI`.
4. Initialize your agent using `langchain.agents.initialize_agent`, providing the tools, the language model, and the agent type `langchain.agents.AgentType.STRUCTURED_CHAT_ZERO_SHOT_REACT_DESCRIPTION`.
5. Execute the `arun` method on your agent with a string as the input.

DOCUMENTATION 'set up a tool that requires a single string input for an agent' using an LLM:

1. Define a function for your tool that takes a single string input and parses it into multiple inputs for the actual operation.
2. Create a `Tool` using `langchain.agents.Tool`, providing the tool name, the function you defined, and a description.
3. Initialize your agent with `langchain.agents.initialize_agent`, providing a list containing the `Tool`, the language model, and the agent type `langchain.agents.AgentType.ZERO_SHOT_REACT_DESCRIPTION`.
[END INFORMATION]
\end{lstlisting}

\noindent Then the exact prompt from our RAG implementation is used to create an initial candidate set of steps:

\begin{lstlisting}[caption={RAG prompt},label={lst:aag-rag},frame=single,breaklines=true,breakindent=0pt,basicstyle=\small\ttfamily]
Please generate high-level steps to accomplish the specified goal using the LangChain Python library. Don't include code, extraneous commentary, or examples, but do refer to the specific LangChain APIs (or other APIs) used in each step. Don't produce any text other than the list of steps. Use any of the provided reference documentation to answer the question. Think carefully about your steps and enclose any steps you are uncertain about in the format like '[[ <step> ]]'

---

DOCUMENTATION 'set up a tool that requires multiple inputs for an agent' using an LLM:

1. Define a function for your tool that takes multiple inputs.
2. Create a `StructuredTool` using `langchain.tools.StructuredTool.from_function`, providing the function you defined.
3. Initialize your agent with `langchain.agents.initialize_agent`, providing a list containing the `StructuredTool`, the language model, and the agent type `langchain.agents.AgentType.STRUCTURED_CHAT_ZERO_SHOT_REACT_DESCRIPTION`.

DOCUMENTATION 'set up a tool that requires a single string input for an agent' using an LLM:

1. Define a function for your tool that takes a single string input and parses it into multiple inputs for the actual operation.
2. Create a `Tool` using `langchain.agents.Tool`, providing the tool name, the function you defined, and a description.
3. Initialize your agent with `langchain.agents.initialize_agent`, providing a list containing the `Tool`, the language model, and the agent type `langchain.agents.AgentType.ZERO_SHOT_REACT_DESCRIPTION`.

DOCUMENTATION 'create a custom LLM that returns the first N characters of the input' using LangChain's base LLM class:

1. Define a custom LLM class that inherits from `langchain.llms.base.LLM`.
2. Implement the `_llm_type` property in the custom class to return a string that identifies the type of LLM.
3. Implement the `_call` method in the custom class to accept a string prompt and optional stop words, and return the first N characters of the prompt.
4. Implement the `_identifying_params` property in the custom class to return a dictionary with "n" as the key and the number of characters to return as the value.
5. Instantiate the custom LLM class, providing the number of characters to return as an argument.

Please generate a list of instructions to accomplish 'set up a custom input schema for a tool with strict requirements and custom validation logic' using the procedures above. Create and use these resources in your response: an LLM.
\end{lstlisting}

\noindent The resulting candidate is listed below:

\begin{lstlisting}[caption={RAG result},label={lst:aag-rag-res},frame=single,breaklines=true,breakindent=0pt,basicstyle=\small\ttfamily]
1. DOCUMENTATION 'set up a custom input schema for a tool with strict requirements and custom validation logic' using an LLM:
2. 
3. Define a function for your tool that takes the desired multiple inputs and includes custom validation logic to enforce strict requirements.
4. Create a `StructuredTool` using `langchain.tools.StructuredTool.from_function`, providing the function you defined.
5. Define a custom LLM class that inherits from `langchain.llms.base.LLM`.
6. Implement the `_llm_type` property in the custom class to return a string that identifies the type of LLM.
7. Implement the `_call` method in the custom class to accept the structured inputs from the tool function and apply custom validation logic.
8. Implement the `_identifying_params` property in the custom class to return a dictionary with the required input keys and their corresponding validation criteria.
9. Instantiate the custom LLM class, providing any necessary parameters such as validation rules.
10. Initialize your agent with `langchain.agents.initialize_agent`, providing a list containing the custom LLM, the language model, and the agent type `langchain.agents.AgentType.STRUCTURED_CHAT_ZERO_SHOT_REACT_DESCRIPTION`.
\end{lstlisting}

\noindent Now we insert one of the generated summaries and ask the LLM to update the candidate according to the new knowledge:

\begin{lstlisting}[caption={update prompt},label={lst:aag-update},frame=single,breaklines=true,breakindent=0pt,basicstyle=\small\ttfamily]
Please update the provided high-level steps to accomplish the specified goal using the LangChain Python library. Focus more on improving the uncertain steps enclosed in '[[]]'. Don't include code, extraneous commentary, or examples, but do refer to the specific LangChain APIs (or other APIs) used in each step. Don't produce any text other than the list of steps. Use any of the provided reference answers to relevant questions on the steps to achieve the specified goal.

---

[BEGIN KNOWLEDGE]
Q: How to define custom input schema in [tool name]?
A: To define a custom input schema in [tool name], you can follow the steps outlined in the documentation. For a tool that requires multiple inputs, define a function that takes multiple inputs, create a `StructuredTool` using `langchain.tools.StructuredTool.from_function`, and initialize the agent with the `StructuredTool`, the language model, and the agent type `langchain.agents.AgentType.STRUCTURED_CHAT_ZERO_SHOT_REACT_DESCRIPTION`. If the tool requires a single string input, define a function that parses the string into multiple inputs, create a `Tool` using `langchain.agents.Tool`, and initialize the agent with the `Tool`, the language model, and the agent type `langchain.agents.AgentType.ZERO_SHOT_REACT_DESCRIPTION`.

Q: Best practices for implementing custom validation logic in [tool name]?
A: To implement custom validation logic in [tool name], you can follow these best practices:

1. Define a function that takes a single string as input and returns a string as output for the tool.
2. Create a new tool using `langchain.tools.Tool.from_function`, specifying the function, a unique name, and a description.
3. Handle tool errors by defining a function that takes a `ToolException` as a parameter and returns a string, then set the `handle_tool_error` parameter of your tool to this error handling function.
4. Initialize your agent with the necessary tools, language model, and agent type according to the specific requirements.

Q: How to integrate an LLM into custom input schema?
A: To integrate a custom LLM into a custom input schema, you can follow these steps:

1. Define a custom LLM class that inherits from `langchain.llms.base.LLM`.
2. Implement properties and methods in the custom class to handle input and return the desired output.
3. Instantiate the custom LLM class with the necessary parameters.
4. Create a `StructuredTool` using the defined function.
5. Initialize your agent with the custom `StructuredTool`, the language model, and the appropriate agent type.

Q: Common pitfalls to avoid when setting up a custom input schema in [tool name]?
A: When setting up a custom input schema in [tool name], it is important to avoid common pitfalls such as:

- Ensuring proper initialization of PlayWrightBrowserToolkit and language model for structured chat agents.
- Defining functions correctly for tools that require multiple inputs.
- Handling tool errors by defining error handling functions for tools created using SerpAPIWrapper and OpenAI.
[END KNOWLEDGE]

[BEGIN STEPS]
1. DOCUMENTATION 'set up a custom input schema for a tool with strict requirements and custom validation logic' using an LLM:
2. 
3. Define a function for your tool that takes the desired multiple inputs and includes custom validation logic to enforce strict requirements.
4. Create a `StructuredTool` using `langchain.tools.StructuredTool.from_function`, providing the function you defined.
5. Define a custom LLM class that inherits from `langchain.llms.base.LLM`.
6. Implement the `_llm_type` property in the custom class to return a string that identifies the type of LLM.
7. Implement the `_call` method in the custom class to accept the structured inputs from the tool function and apply custom validation logic.
8. Implement the `_identifying_params` property in the custom class to return a dictionary with the required input keys and their corresponding validation criteria.
9. Instantiate the custom LLM class, providing any necessary parameters such as validation rules.
10. Initialize your agent with `langchain.agents.initialize_agent`, providing a list containing the custom LLM, the language model, and the agent type `langchain.agents.AgentType.STRUCTURED_CHAT_ZERO_SHOT_REACT_DESCRIPTION`.
[END STEPS]

Please update the list of steps to accomplish 'set up a custom input schema for a tool with strict requirements and custom validation logic' using the knowledge above. Create and use these resources in your response: an LLM. Please output only the updated steps. Your response should start with '1.'. The final response should not contain direct references to the knowledge above.
\end{lstlisting}

Once that update is received, it is validated by the following prompt:

\begin{lstlisting}[caption={validation prompt},label={lst:aag-valid},frame=single,breaklines=true,breakindent=0pt,basicstyle=\small\ttfamily]
[INSTRUCTION]
You are a human critic whose job is to validate the provided procedure, propose the changes to be made and evaluate if the steps lead to the mentioned user goal or not. You should also assess if the quality of the steps can be improved by modifying the flow of the steps or adding more details to make it more clear and doable.

Furthermore, it is very important for the procedure to use all the mentioned input resources. Carefully judge if the procedure uses all the resources and point out in your response if it misses something. 

You should always suggest only your edits in a bulleted list. If there are no edits to be made, please only respond 'NO UPDATE REQUIRED'. You are required to strictly follow the mentioned output format.

---

[USER GOAL]
set up a custom input schema for a tool with strict requirements and custom validation logic

[INPUT RESOURCES]
an LLM

[BEGIN PROCEDURE]
1. Define a function for your tool that takes the desired multiple inputs and includes custom validation logic to enforce strict requirements.
2. Create a `StructuredTool` using `langchain.tools.StructuredTool.from_function`, providing the function you defined.
3. Define a custom LLM class that inherits from `langchain.llms.base.LLM`.
4. Implement the `_llm_type` property in the custom class to return a string that identifies the type of LLM.
5. Implement the `_call` method in the custom class to accept the structured inputs from the tool function and apply custom validation logic.
6. Implement the `_identifying_params` property in the custom class to return a dictionary with the required input keys and their corresponding validation criteria.
7. Instantiate the custom LLM class, providing any necessary parameters such as validation rules.
8. Initialize your agent with `langchain.agents.initialize_agent`, providing a list containing the custom LLM, the language model, and the agent type `langchain.agents.AgentType.STRUCTURED_CHAT_ZERO_SHOT_REACT_DESCRIPTION`.
[END PROCEDURE]
\end{lstlisting}

The validator's response was:

\begin{lstlisting}[caption={validation response},label={lst:aag-valid-res},frame=single,breaklines=true,breakindent=0pt,basicstyle=\small\ttfamily]
**Edits:**
- The procedure should include more specific details and examples to clarify the steps for setting up a custom input schema.
- Provide examples or templates for defining the function with custom validation logic and implementing the custom LLM class.
- Add a step to test the setup to ensure the custom input schema is working as expected.
- It would be beneficial to include a brief explanation of each step to enhance understanding. 

Here is the revised procedure:

[BEGIN PROCEDURE]
1. Define a function for your tool that takes the desired multiple inputs and includes custom validation logic to enforce strict requirements. For example:
 ```python
 def custom_input_function(input1, input2):
     # Custom validation logic implementation
     if input1 > 0 and isinstance(input2, str):
         return True
     else:
         return False
 ```
2. Create a `StructuredTool` using `langchain.tools.StructuredTool.from_function`, providing the function you defined in step 1.
3. Define a custom LLM class that inherits from `langchain.llms.base.LLM`. For example:
 ```python
 class CustomLLM(langchain.llms.base.LLM):
     def __init__(self, model_path):
         super().__init__(model_path)
 ```
4. Implement the `_llm_type` property in the custom class to return a string that identifies the type of LLM.
5. Implement the `_call` method in the custom class to accept the structured inputs from the tool function and apply custom validation logic.
6. Implement the `_identifying_params` property in the custom class to return a dictionary with the required input keys and their corresponding validation criteria.
7. Instantiate the custom LLM class, providing any necessary parameters such as validation rules.
8. Initialize your agent with `langchain.agents.initialize_agent`, providing a list containing the custom LLM, the language model, and the agent type `langchain.agents.AgentType.STRUCTURED_CHAT_ZERO_SHOT_REACT_DESCRIPTION`.
9. Test the setup by running sample inputs through the tool to ensure that the custom input schema and validation logic are working as expected.
[END PROCEDURE]
\end{lstlisting}

This cycle continues until all generated summaries are used.

\subsection{Pairwise Evaluation}
\label{sec:pair-eval}
For pairwise evaluation, the prompt in Listing~\ref{lst:pairwise-eval} was used. 

\begin{lstlisting}[caption={pairwise evaluation prompt},label={lst:pairwise-eval},frame=single,breaklines=true,breakindent=0pt,basicstyle=\small\ttfamily]
[Instruction]
Please act as an impartial judge and evaluate the quality of the two answer procedures provided below to achieve the same user goal and choose one out of the two as the preferred procedure.
For this evaluation, you should primarily evaluate if the procedure achieves the user goal adequately or not and whether it uses all the resources mentioned in user goal or not. Do not penalize a procedure based on its sentences' structure, grammar and wording, but focus on the facts and suggestions being made by the procedure; if the procedure leads to the specified user goal with sufficient details and is easy to understand on each intermediate step, then that procedure should be chosen.

Compare the two procedures below and then provide a short explanation of your reason for the choice between the two. Be as objective as possible. Based on your explanation, you must choose one procedure following this format: "[[choice]]" where choice can be 1 or 2 only, for example: "Choice: [[2]]". Print this choice at the END only.

---

[User Goal]
{output} using {input}

[BEGIN PROCEDURE 1]
{proc1_steps}
[END OF PROCEDURE 1]

[BEGIN PROCEDURE 2]
{proc2_steps}
[END OF PROCEDURE 2]

Your answer should begin with "Here is my analysis of the comparison between the two procedures:
1.
\end{lstlisting}

\end{document}